\documentclass[letterpaper, 10 pt, conference]{ieeeconf}  

\IEEEoverridecommandlockouts                              

\usepackage{amssymb}  

\usepackage{microtype}
\usepackage{graphicx}
\usepackage{booktabs} 

\usepackage{hyperref}

\usepackage{amssymb}
\usepackage{amsmath}
\usepackage{mathtools}

\usepackage[capitalize,noabbrev]{cleveref}

\usepackage{caption}
\usepackage{subcaption}

\usepackage{xcolor}

\usepackage[]{mdframed}

\title{Weakly-supervised VLM-guided Partial Contrastive Learning for Visual Language Navigation}

\author{Ruoyu Wang$^{1}$, Tong Yu$^{2}$, Junda Wu$^{3}$, Yao Liu$^{4}$, Julian McAuley$^{5}$, Lina Yao$^{6}$ 
\thanks{$^{1}$ Ruoyu Wang, Corresponding Author, University of New South Wales, Sydney, Australia
        {\tt\small ruoyu.wang5@unsw.edu.au}}%
\thanks{$^{2}$ Tong Yu, Adobe Research, San Jose, United States
        }%
\thanks{$^{3}$ Junda Wu, University of California San Diego, La Jolla, United States
        }%
\thanks{$^{4}$ Yao Liu, Northeastern University, China
        }%
\thanks{$^{5}$ Julian McAuley, University of California San Diego, La Jolla, United States
        }%
\thanks{$^{6}$ Lina Yao, CSIRO's Data61 and University of New South Wales, Australia
        }%
}

\begin{document}

\maketitle
\thispagestyle{empty}
\pagestyle{empty}

\begin{abstract}

Visual Language Navigation (VLN) is a fundamental task within the field of Embodied AI, focusing on the ability of agents to navigate complex environments based on natural language instructions. Despite the progress made by existing methods, these methods often present some common challenges. First, they rely on pre-trained backbone models for visual perception, which struggle with the dynamic viewpoints in VLN scenarios. Second, the performance is limited when using pre-trained LLMs or VLMs without fine-tuning, due to the absence of VLN domain knowledge. Third, while fine-tuning LLMs and VLMs can improve results, their computational costs are higher than those without fine-tuning. To address these limitations, we propose Weakly-supervised Partial Contrastive Learning (WPCL), a method that enhances an agent's ability to identify objects from dynamic viewpoints in VLN scenarios by effectively integrating pre-trained VLM knowledge into the perception process, without requiring VLM fine-tuning. Our method enhances the agent's ability to interpret and respond to environmental cues while ensuring computational efficiency. Experimental results have shown that our method outperforms the baseline methods on multiple benchmarks, which validate the effectiveness, robustness and generalizability of our method.

\end{abstract}

\section{Introduction}
\label{intro}

\begin{figure}
    \centering
    \includegraphics[width=0.8\linewidth]{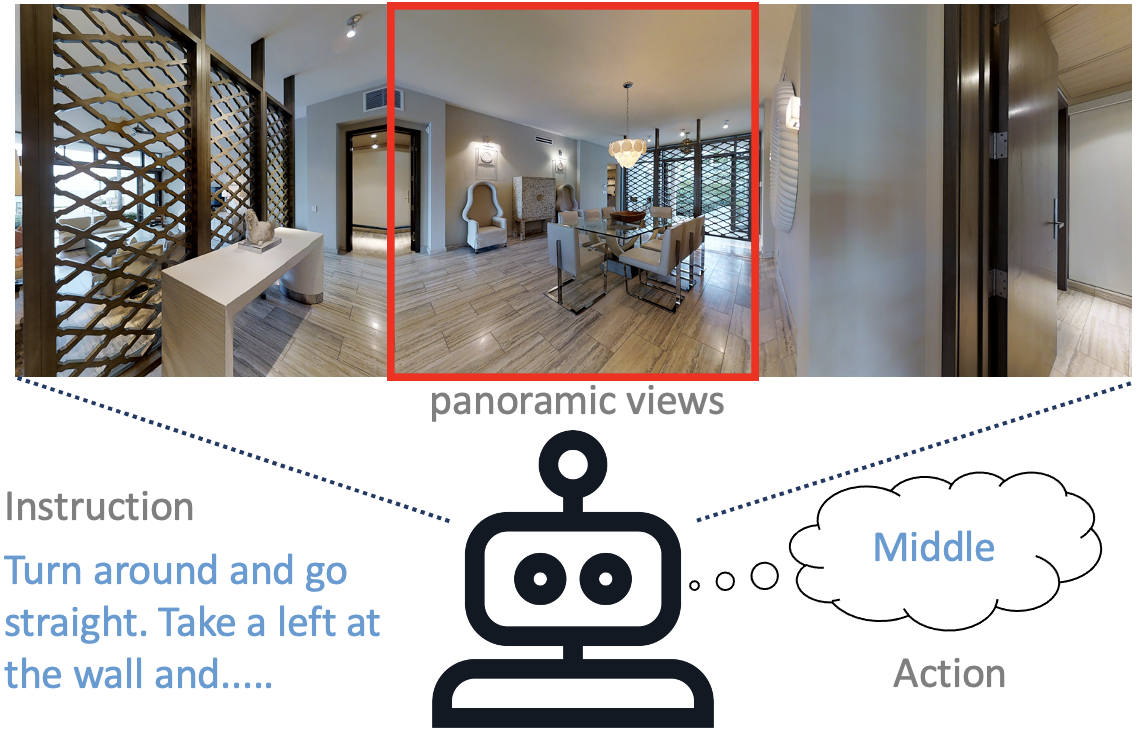}
    \caption{An example of Visual Language Navigation task. The agent follows a natural language instruction and navigates from an initial viewpoint within the given environment. At each time step, the agent chooses one view from its surrounding panoramic options to move towards.}
    \label{fig:vln}
    \vspace{-10pt}
\end{figure}

Vision-Language Navigation (VLN) is a key research area in Embodied AI, enabling agents to navigate based on human instructions (Figure~\ref{fig:vln}). It has gained increasing attention in robotics and machine learning \cite{gu2022vision}. Over the years, various methods have been developed to advance VLN. These methods usually involve model training steps, some are trained in an end-to-end manner \cite{anderson2018vision,fried2018speaker,tan2019learning}, and some incorporate VLN-specific pre-training and fine-tuning steps \cite{hao2020towards,hong2020recurrent,wang2023scaling}. Recently, the rise of pre-trained Large Language Models (LLMs) and Vision-Language Models (VLMs) has fueled interest in leveraging these models for VLN. Studies show that effective agents can be produced with minimal training \cite{zhou2024navgpt,chen2024mapgpt,long2024discuss} by integrating the pre-trained knowledge into VLN systems. Some approaches also fine-tune LLMs for VLN-specific tasks \cite{lin2024navcot,pan2023langnav,zheng2024towards}, typically by presenting an image and asking the model what action to take next.

However, these methods often present some common challenges. First, they often use pre-trained backbone models such as ResNet \cite{he2016deep} or CLIP \cite{radford2021learning} for visual perception, which is not ideally suited for the scenario of VLN due to the constant change in viewpoint \cite{fan2024active} (Section~\ref{background_intuition}). Second, when incorporating pre-trained LLMs or VLMs without any fine-tuning, the method often results in performance that falls short of traditional methods due to the lack of VLN-specific information \cite{zhou2024navgpt,chen2024mapgpt,long2024discuss}. Third, while fine-tuning LLMs or VLMs with Vision-Language Navigation data can enhance performance, it is often computationally expensive \cite{lin2024navcot,pan2023langnav,zheng2024towards} and therefore not always feasible and affordable.

To overcome these limitations, we propose Weakly-supervised Partial Contrastive Learning (WPCL), which: (1) enhances environmental understanding by identifying objects from dynamic viewpoints, addressing shortcomings of pre-trained visual perception backbones; (2) introduces a weakly-supervised mechanism to reduce reliance on LLM/VLM outputs, mitigating biases while leveraging their zero-shot capabilities; and (3) integrates pre-trained large VLMs into VLN without fine-tuning, avoiding associated computational costs. In summary, our contributions are threefold:

\begin{itemize}
    \item We introduce a novel causal perspective to understand VLN tasks, analyze the association structure within the observation history and design a partial contrastive learning framework tailored to the scenario.
    \item We present a novel learning paradigm tailored to the VLN task, which integrates the knowledge from the pre-trained large foundation models into VLN and effectively mitigates biases inherent in pre-trained models.
    \item We propose a novel framework for Vision-Language Navigation and provide empirical evidence of its effectiveness across various benchmarks. Comprehensive analysis further demonstrates the robustness of our method across different pre-trained VLMs.
\end{itemize}

\begin{figure}
    \centering
    \includegraphics[width=0.9\linewidth]{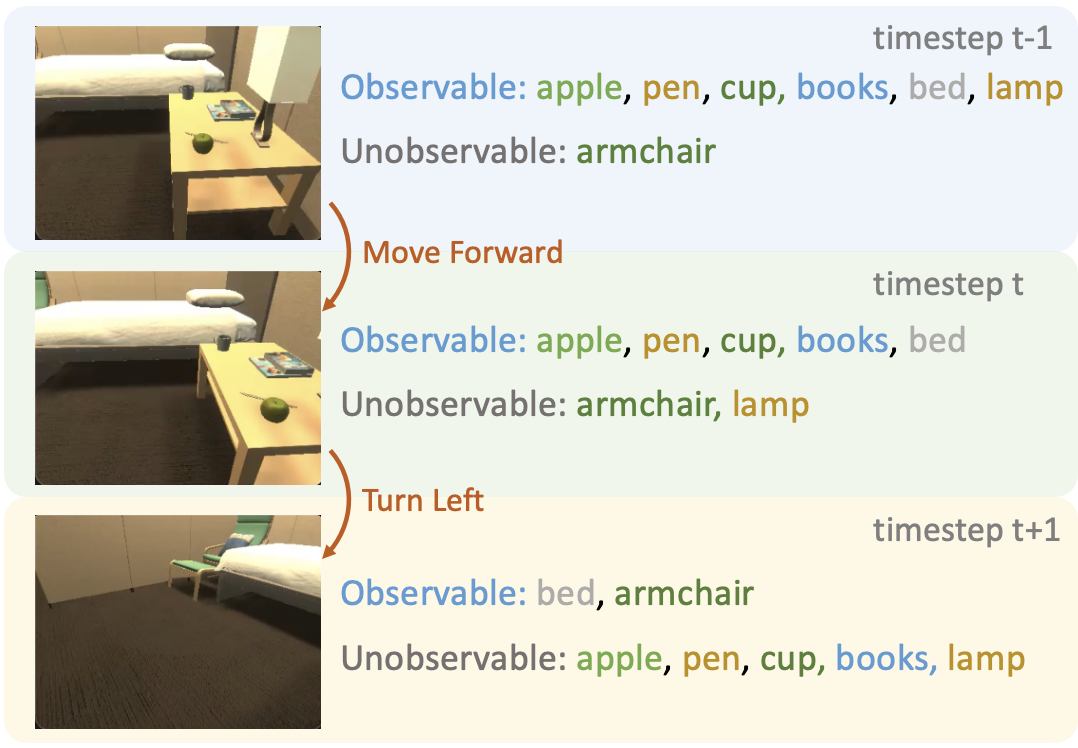}
    \caption{
    Illustrative example of our motivation. The agent's actions do not alter the environment but only change its viewpoint, resulting in different observations. The agent should recognize the same object from different angles by identifying common elements across observations.}
    \label{fig:partial_contrast}
    \vspace{-10pt}
\end{figure}

\section{Background}

\subsection{Motivation}
\label{background_intuition}
In Navigation, an agent's actions do not alter the environment but only shift its viewpoint, changing the perspective from which objects are observed (or unobserved). For example, in Figure~\ref{fig:partial_contrast}, when an agent moves from timestep $t-1$ to $t$ via \textit{Move Forward}, most objects remain visible from a different angle, except for the lamp, which disappears due to the viewpoint change. After \textit{Turn Left} at $t+1$, the armchair enters view while objects like the books become unobservable. Although all objects persist in the environment, the agent’s viewpoint dictates their observability.

Given these viewpoint shifts, \textbf{an ideal agent with a comprehensive understanding of the environment should recognize the same objects from different angles}. However, many VLN methods rely on pre-trained vision models like ResNet~\cite{he2016deep} or CLIP~\cite{radford2021learning}, which have been shown to be highly sensitive to viewpoint changes, making it difficult to consistently recognize the same object from different perspectives \cite{fan2024active}, limiting overall VLN performance.

\begin{figure}
    \centering
    \includegraphics[width=0.8\linewidth]{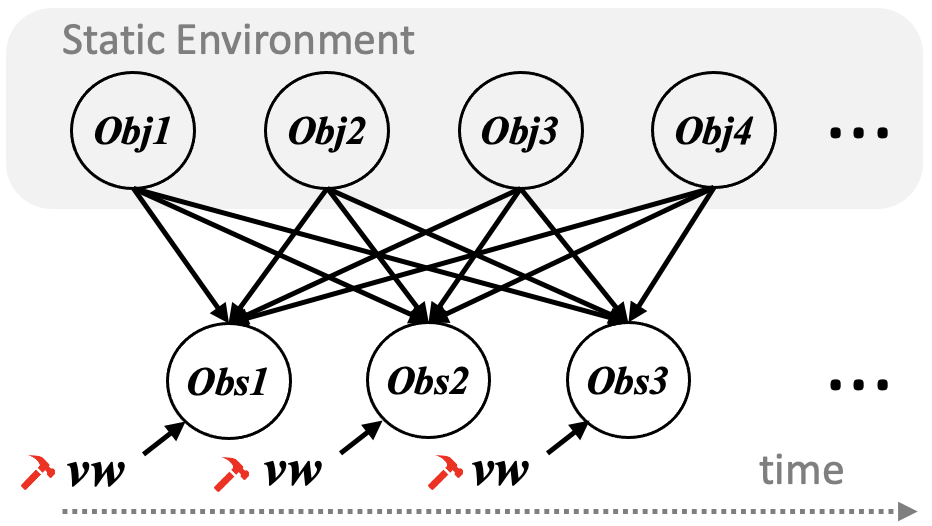}
    \caption{A generalized causal framework between objects and observations. As the agent moves, its changing viewpoint alters the way objects appear in its observations.}
    \label{fig:causal_graph}
    \vspace{-10pt}
\end{figure}

To address this, we introduce a generalizable causal framework (Figure~\ref{fig:causal_graph}), where $Obs$ represents the agent's observations, $Obj$ denotes the objects in the environment, and $vw$ stands for the viewpoint. The $Obj$ layer remains static over time, and the $Obj$ and $Obs$ layers are fully connected, indicating that all objects can potentially appear in an observation, depending on the viewpoint. At each timestep, the agent's action induces an intervention on the viewpoint $vw$, which does not causally affect the environment but influences how the agent perceives it, thus determining how objects are represented in that observation.

Within this causal framework, objects act as confounders between observations, especially when two observations share the same object. To enable the agent to consistently identify the same object from various viewpoints, we employ a contrastive learning framework to encourage the agent to extract the mutual information (MI) between these observations.

\subsection{Why Contrastive Learning}
\label{background_contrastive}
The framework in Section~\ref{background_intuition} reveals the data generation process of an agent's observations. And Contrastive Learning can be leveraged to encourage the agent better understand the environment as it's shown to be effective on inverting this process by extract mutual information (MI) between the positive samples \cite{zimmermann2021contrastive}. A typical contrastive learning framework generates positive samples through random augmentations \cite{chen2020simple,chen2020improved} and an encoder is trained to bring them closer in the latent space by minimizing the InfoNCE loss:
\begin{equation}
    \mathcal{L} = \mathbb{E}_{v \in D} \left[ -\log \frac{\mathcal{Q}(v_{1}^{+},v_{2}^{+})}{\mathcal{Q}(v_{1}^{+},v_{2}^{+}) + \sum_{k=1}^{K} \mathcal{Q}(v_{1}^{+},v_{k}^{-})} \right]
  \label{eq:infonce}
\end{equation}
\begin{equation}
    \mathcal{Q}(v_{i},v_{j}) = \exp(\frac{z_{i}^{\top}z_{j}}{\tau \cdot \rVert z_{i} \rVert\rVert z_{j} \rVert})
  \label{eq:infonce_q_pos}
\end{equation}
where $v_{1}^{+}$ and $v_{2}^{+}$ denotes a pair of positive samples, $v_{k}^{-}$ denotes a negative sample, $z_{i}$ denotes the feature vector of an image, and $\tau$ is the temperature parameter.

To adapt this framework to our scenario, we select pairs of images from the observation history that contain common objects as positive samples. This allows the framework to extract mutual information (MI) (i.e., the common objects) between these samples, inverting the data generation process and enhancing the agent's understanding of the environment.

\subsection{Why weakly-supervised by VLM}
\label{background_weakly_supervised}

We choose a VLM-guided weakly-supervised mechanism for two key reasons. First, the Contrastive Learning framework introduced in Section~\ref{background_contrastive} is not directly applicable to our scenario, as it assumes that positive samples share mutual information. This assumption is usually valid when positive samples are generated by image augmentation \cite{chen2020simple}. However, in our case, randomly selecting two observations from the observation history does not guarantee that they will have mutual information. Therefore, we leverage the zero-shot capability of VLM to guide the selection of positive samples.

Second, as discussed in Section~\ref{intro}, methods relying heavily on pre-trained LLMs or VLMs often underperform VLN-tailored training due to lacking task-specific information. Therefore, we minimize reliance on pre-trained models by using a weakly-supervised approach instead of a fully-supervised one, potentially enhancing the performance while circumventing the computational costs of VLM fine-tuning.

\begin{figure*}
  \centering
  \includegraphics[width=0.95\linewidth]{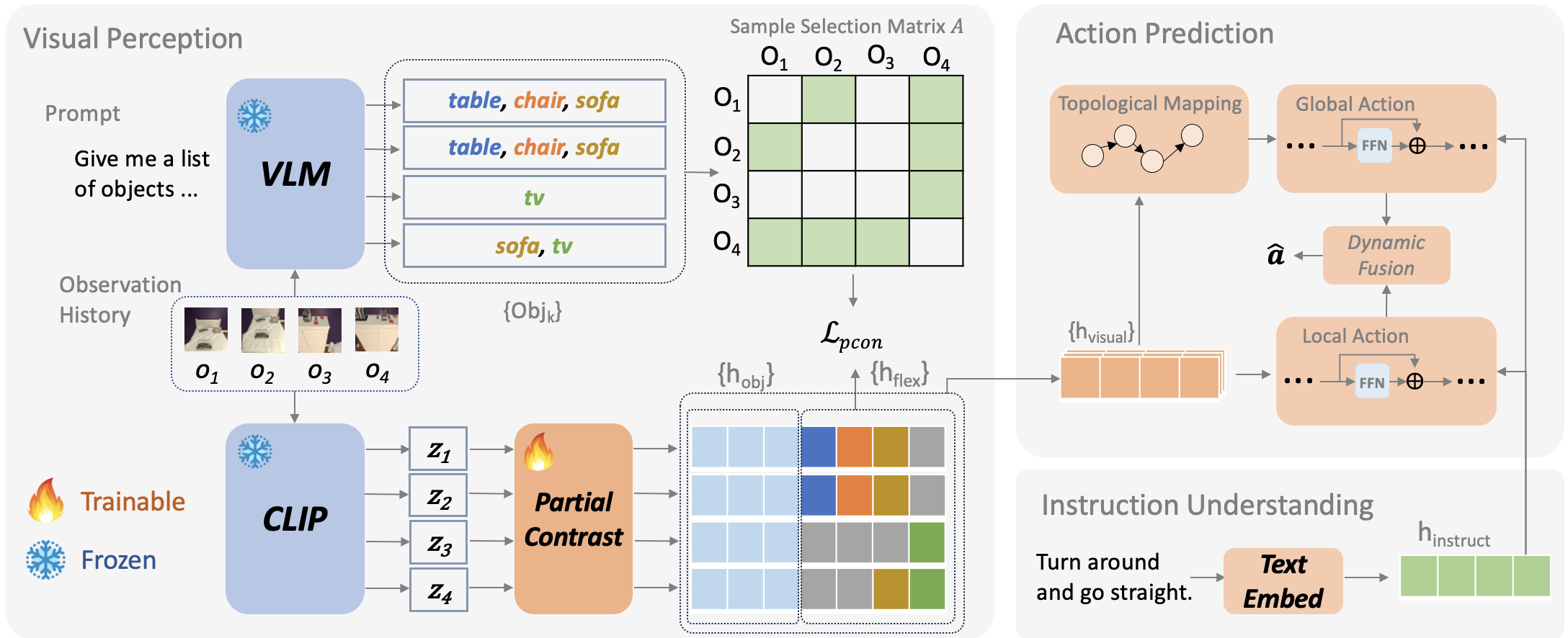}
  \caption{The overall workflow of our method. The observation histories are first processed by the pre-trained CLIP model. And a VLM is used to generate the Sample Selection Matrix $A$ (Section~\ref{method_supervised_vlm}), which guides the weakly-supervised contrastive learning framework to refine the visual features obtained from the pre-trained visual backbone (Section~\ref{method_partial_contra}). The final visual features are subsequently integrated into the DUET framework for VLN.}
  \label{fig:framework}
  \vspace{-10pt}
\end{figure*}

\section{Our method}
\label{method}

\subsection{Weakly-supervised by pre-trained VLM}
\label{method_supervised_vlm}

As elaborated in Section~\ref{background_weakly_supervised}, we leverage the zero-shot capability of pre-trained Vision-Language Models (VLMs) to assist in positive sample selection. Specifically, given an observation history $O = \{o_{i}\}_{i=1}^{N}$ where $N$ is the number of historical observations, we independently input each $o_{i}$ into the pre-trained LLaVA v1.5 \cite{liu2023llava}, asking it to identify objects in the image using the following prompt to extract a list of objects contained in each image:

\begin{mdframed}
Give me a list of objects that appear in the given image. Ignore the general environmental objects such as walls, floors or rooftops. Please only answer with a list of objects, and separate each object by a comma.
\end{mdframed}

We specified the expected output format in the prompt to ensure consistency across all images in the observation history. The same prompt and output templates were applied uniformly. Below is a sample output:

\begin{mdframed}
desk, chair, television, sofa, coffee table, fruits
\end{mdframed}

We process this output sentence by splitting it at commas to obtain a set of objects 
$\{ Obj_{k} \}^{(i)}$ for each observation $o_{i}$. Further, we construct a sample selection matrix $A$, which is an $N \times N$ symmetric binary matrix, where $A_{i,j} = A_{j,i} = 1$ if and only if $|\{ Obj_{k} \}^{(i)} \cap \{ Obj_{k} \}^{(j)}| \geq m$; otherwise, $A_{i,j} = A_{j,i} = 0$, where $m$ is a tunable hyper-parameter.

We incorporate this sample selection matrix $A$ as the weak supervisory signal into the agent's training process, which will be elaborated in Section~\ref{method_partial_contra}.

\subsection{Partial Contrastive Learning}
\label{method_partial_contra}

We employed a Partial Contrastive Learning framework to encourage the agent to recognise objects from different angles as it navigates the environment. First, with a given set of observation history $O=\{o_{1}, o_{2}, ..., o_{n}\}$, we obtain the sample selection matrix $A$ with the methods in Section~\ref{method_supervised_vlm}. Then we define observation $(o_{i},o_{j})$ as a positive sample pair if and only if $A_{i,j} = A_{j,i} = 1$. Formally, the set of all positive sample pairs $\mathcal{P}$ is selected by the following condition:
\begin{equation}
    (o_{i}, o_{j}) \in \mathcal{P} \iff A_{i,j} = 1
    \label{eq:positive_sample}
\end{equation}

Further, for any positive sample pair $(o_{i}, o_{j})$, we consider an observation $o_{k}$ as a negative sample if and only if $o_{k}$ does not have more than $m$ common objects with both $o_{i}$ and $o_{j}$:
\begin{equation}
    \left( A_{i,k} = 0 \land A_{j,k} = 0 \right) \implies o_{k} \in \mathcal{N}
    \label{eq:negative_sample}
\end{equation}
where $\mathcal{N}$ denotes the set of negative samples. Then, unlike the typical contrastive learning framework introduced in Section~\ref{background_contrastive}, which assumes the entire representation between a pair of positive samples should be fully invariant, we relax such assumptions. We assume that only a portion of the representation should remain invariant across a pair of positive samples $o_{i}$ and $o_{j}$, which is used to denote the common objects in $\{ Obj_{k} \}^{(i)} \cap \{ Obj_{k} \}^{(j)}$. The rest of the representations are not necessarily the same, as they may contain other information in the image, including but not limited to noises, information related to the viewpoint, and those un-shared objects in $\left( \{ Obj_{k} \}^{(i)} \setminus \{ Obj_{k} \}^{(j)} \right) \cup \left( \{ Obj_{k} \}^{(j)} \setminus \{ Obj_{k} \}^{(i)} \right)$.

Formally, we introduce a tunable parameter $ 0 < \lambda < 1$ denoting the portion of features we use for the partial contrastive learning framework, and divide the visual input into two separate segments as shown in Equation~\ref{eq:feature_split}:
\begin{equation}
    h_{visual} = \left[ h_{objs}; h_{flex} \right]
    \label{eq:feature_split}
\end{equation}
where $h_{visual} \in \mathbb{R}^{d}$ represents the $d$-dimensional visual feature of the observation. A segment of $h_{visual}$, denoted as $h_{objs} \in \mathbb{R}^{\lambda d}$, is allocated to store information about objects in the intersection $\{ Obj_{k} \}^{(i)} \cap \{ Obj_{k} \}^{(j)}$ within the observation. The remaining part, $h_{flex} \in \mathbb{R}^{(1-\lambda) d}$, is designated for encoding other information present in the observation.

Finally, as elaborated Section~\ref{background_contrastive}, with the selected positive and negative samples, we applied the InfoNCE Loss only on the $h_{objs}$, and implemented a partial InfoNCE Loss as demonstrated in Equation~\ref{eq:infonce_partial} - Equation~\ref{eq:infonce_q_pos_partial}.
\begin{equation}
    \mathcal{L}_{pcon} = \mathbb{E}_{v \in D} \left[ -\log \frac{\mathcal{Q}(o_{i},o_{j})}{\mathcal{Q}(o_{i},o_{j}) + \sum_{k=1}^{K} \mathcal{Q}(o_{i},o_{k})} \right]
  \label{eq:infonce_partial}
\end{equation}
\begin{equation}
    \mathcal{Q}(v^{(1)},v^{(2)}) = \exp \left\{ \frac{h_{obj}^{(1)\,\top} h_{obj}^{(2)}}{\tau \cdot \| h_{obj}^{(1)} \| \| h_{obj}^{(2)} \|} \right\}
  \label{eq:infonce_q_pos_partial}
\end{equation}
where $o_{i}$ and $o_{j}$ denotes a pair of positive samples from set $\mathcal{P}$, $o_{k}$ denotes a negative sample from set $\mathcal{N}$, $v^{(1)}$ and $v^{(2)}$ denotes any images either positive or negative, $h_{obj}^{(1)}$ and $h_{obj}^{(2)}$ denotes the object feature vector of $v^{(1)}$ and $v^{(2)}$, $\tau$ is the temperature parameter. This loss consequently encourages the model to identify the same objects from different angles across the positive samples.

\subsection{Overall Workflow}
\label{method_ours}

The overall workflow of our proposed method is depicted in Figure~\ref{fig:framework}, which consists of three key modules: Visual Perception, Instruction Understanding, and Action Prediction. Our primary contribution lies in the Visual Perception module.

First, a set of observation history $\{o_{i}\}_{i=1}^{N}$ is fed into the pre-trained CLIP \cite{radford2021learning} to generate a set of image features $\{z_{i}\}_{i=1}^{N}$. We use the pre-trained CLIP for image processing due to its proven effectiveness in navigation tasks \cite{khandelwal2022simple} and its low computational resource requirements. Then, we pass these features into a trainable Partial Contrast Module to produce a new set of image features $h_{visual} = \{h_{i}\}_{i=1}^{N}$, which are segmented into two pieces $\{h_{obj}\}_{i=1}^{N}$ and $\{h_{flex}\}_{i=1}^{N}$ as shown in Equation~\ref{eq:feature_split}. Meanwhile, the observation history $\{o_{i}\}_{i=1}^{N}$ is fed into the pre-trained VLM to generate a list of items $\{ Obj_{k} \}^{(i)}$ for each observation $o_{i}$, which is then used to construct the sample selection matrix $A$ as supervisory signals to guide the agent to be able to identify objects from the changing viewpoints, as introduced in Section~\ref{method_supervised_vlm}. 

In the Instruction Understanding Module, we use a text embedding layer to convert the input instruction into text features $h_{instrct}$. Then, both $h_{instrct}$ and the visual features $h_{visual}$ are fed into the Action Prediction Module. We adapt the architecture of Dual-Scale Graph Transformer (DUET) \cite{chen2022think} in the Action Prediction Module, which consists of a Global Action prediction Module incorporated with a Topological Mapping algorithm, and a Local Action prediction Module to predict action solely based on the visual and instruction input. Finally, a Dynamic Fusion module is used to consolidate the Global and Local Action prediction to produce the final prediction of the action $\hat{a}$.

During the training stage, following the pipeline in existing works \cite{chen2022think,chen2021history,hao2020towards}, we first pre-trained the models based on off-line expert demonstrations with single-step action prediction (SAP) \cite{chen2021history} and object grounding (OG) \cite{lin2021scene} if object annotations are available:
\begin{equation}
    \mathcal{L}_{SAP} = \sum_{t=1}^{T} -\log p(a_{t}^{*} | W, P_{<t}^{*})
    \label{eq:sap_loss}
\end{equation}
\begin{equation}
    \mathcal{L}_{OG} = -\log p(o^{*} | W, P_{T})
    \label{eq:og_loss}
\end{equation}
where $a^{*}_{t}$ is the expert action of a partial demonstration path $P_{<t}^{*}$, and $o^{*}$ is the groundtruth object at the last location $P_{T}$. 
We further train the models with the supervision from a pseudo interactive demonstrator (PID): 
\begin{equation}
    \mathcal{L}_{PID} = \sum_{t=1}^{T} -\log p(a_{t}^{\pi{*}} | W, P_{<t})
    \label{eq:pid_loss}
\end{equation}
where $a_{t}^{\pi{*}}$ is our pseudo target at step t. Together with the Partial-Contrastive Objective proposed in Equation~\ref{eq:infonce_partial} and Equation~\ref{eq:infonce_q_pos_partial}, the overall objective of our method is:
\begin{equation}
    \min_{\theta} \gamma \mathcal{L}_{SAP} + \mathcal{L}_{PID} + \mathcal{L}_{OG} + \alpha \mathcal{L}_{pcon}
    \label{eq:total_loss}
\end{equation}
where $\gamma$ and $\alpha$ are the weight parameters that regulate the contribution of the corresponding loss terms, ensuring a balanced optimization process.

\begin{table*}[t]
  \centering
  \small
  \caption{
  Results on the R2R benchmark. The \textbf{top section} lists methods that do not utilize any pre-trained large models, the \textbf{middle section} lists methods that involve LLM fine-tuning, and the \textbf{bottom section} lists methods that incorporate pre-trained large models. \textbf{Our method} belongs to the bottom section and outperforms not only other methods in this category but also those in the other sections. A \textbf{-} indicates unavailable performance data for that metric in the corresponding paper.
  }
  \setlength{\tabcolsep}{5pt}
  \renewcommand{\arraystretch}{1}
  \begin{tabular}{l | c c c c c | c c c c c}
    \toprule
    & \multicolumn{5}{c}{Val Seen} & \multicolumn{5}{c}{Val Unseen} \\
    & TL & NE $\downarrow$ & OSR $\uparrow$ & SR $\uparrow$ & SPL $\uparrow$ & TL & NE $\downarrow$ & OSR $\uparrow$ & SR $\uparrow$ & SPL $\uparrow$ \\
    
    \midrule
    
    Seq2Seq \cite{anderson2018vision} & 11.33 & 6.01 & 53 & 39 & - & 8.39 & 7.81 & 28 & 21 & - \\
    Speaker Follower \cite{fried2018speaker} & - & 3.36 & 74 & 66 & - & - & 6.62 & 45 & 36 & - \\
    EnvDrop \cite{tan2019learning} & 11.00 & 3.99 & - & 62 & 59 & 10.70 & 5.22 & - & 52 & 48 \\

    PREVALENT \cite{hao2020towards} & 10.32 & 3.67 & - & 69 & 65 & 10.19 & 4.71 & - & 58 & 53 \\
    VLN$\circlearrowright$BERT \cite{hong2020recurrent} & 11.13 & 2.90 & - & 72 & 68 & 12.01 & 3.93 & - & 63 & 57 \\
    HAMT \cite{chen2021history} & 11.15 & 2.51 & - & 76 & 72 & 11.46 & \textbf{2.29} & - & 66 & 61 \\
    BEVBert \cite{an2022bevbert}& 13.56 & 2.17 & 88 & 81 & 74 & 14.55 & 2.81 & 84 & 75 & 64 \\
    DUET \cite{chen2022think} & 12.32 & 2.28 & 86 & 79 & 73 & 13.94 & 3.31 & 81 & 72 & 60 \\
    GOAT \cite{wang2024vision} & - & \textbf{1.79} & 88 & \textbf{83} & 79 & - & 2.40 & 84 & 77 & 68 \\
    GridMM \cite{wang2023gridmm} & - & - & - & - & - & - & 2.83 & - & 75 & 64 \\

    \midrule
    
    NavCoT \cite{lin2024navcot} & 10.08 & 6.46 & 48 & 41 & 38 & 9.95 & 6.26 & 48 & 40 & 37 \\
    LangNav \cite{pan2023langnav} & - & - & - & 56 & - & - & - & - & 46 & - \\
    NaviLLM \cite{zheng2024towards} & - & - & - & - & - & 12.81 & 3.51 & - & 67 & 59 \\
    NavGPT-2 \cite{zhou2024navgpt2} & 13.08 & 2.98 & 79 & 74 & 65 & 13.25 & 3.18 & 80 & 71 & 60 \\
    
    \midrule
    
    NavGPT \cite{zhou2024navgpt} & - & - & - & - & - & 11.45 & 6.46 & 42 & 34 & 29\\
    MapGPT \cite{chen2024mapgpt} & - & - & - & - & - & - & 6.92 & 58 & 39 & 26 \\
    DicussNav \cite{long2024discuss} & - & - & - & - & - & 9.69 & 5.32 & 61 & 43 & 40 \\

    \midrule
    
    WPCL (Ours) & 11.26 & 1.91 & \textbf{89} & 82 & \textbf{80} & 12.83 & 2.69 & \textbf{86} & \textbf{78} & \textbf{70} \\
    
    \bottomrule
  \end{tabular}
  \label{tab:rst_r2r}
  \vspace{-10pt}
\end{table*}

\section{Experiments}
\label{exp}

\subsection{Experimental Settings}
\label{exp_setting}

\subsubsection{Datasets}
We evaluate our method on three widely used VLN benchmarks: R2R \cite{anderson2018vision}, REVERIE \cite{qi2020reverie} and SOON \cite{zhu2021soon}. \textbf{R2R} provides step-by-step instructions that guide the agent to navigate to a specified location. On average, the instructions consist of 32 words, and the expert paths comprise 6 steps. An example is given in Figure~\ref{fig:vln}. \textbf{REVERIE} provides high-level instructions that describe target locations and objects. On average, the instructions consist of 21 words. The length of expert paths ranges from 4 to 7 steps. \textbf{SOON} provides instructions that describe target rooms and objects, with an average instruction length of 47 words. The length of expert paths varies from 2 to 21 steps, with an average of 9.5 steps. Following standard evaluation conventions, we assess methods in both seen and unseen environments. The val \textbf{seen} split includes training environments, while val \textbf{unseen} tests generalization to novel scenes.

\subsubsection{Baselines}
The baseline methods we compare against are listed in each results table across the three datasets. These methods vary across datasets based on their availability in corresponding papers. Our primary focus is on R2R, the most popular benchmark, as it provides the most widely available results for our comparison. We broadly categorize existing methods into three groups: those without pre-trained large models, those using pre-trained large models with fine-tuning, and those using pre-trained large models without fine-tuning. \textbf{Our method} leverages the knowledge from the pre-trained VLM without fine-tuning it, thus lies in the \textbf{third category} and directly comparable to the methods in that category. 


\textbf{Methods without LLMs/VLMs} This type of method does not leverage any pre-trained knowledge in the foundation models. Some can be trained in an E2E manner such as Seq2Seq \cite{anderson2018vision}, RCM \cite{wang2019reinforced}, Speaker-Follower \cite{fried2018speaker}, and EnvDrop \cite{tan2019learning}, and some others involves VLN-specific pre-training and fine-tuning steps, such as PREVALENT \cite{hao2020towards}, VLN$\circlearrowright$BERT \cite{hong2020recurrent}, HAMT \cite{chen2021history}, BEVBert \cite{an2022bevbert}, GridMM\cite{wang2023gridmm} DUET \cite{chen2022think}, and GOAT \cite{wang2024vision}. While ScaleVLN \cite{wang2023scaling} and MiniVLN \cite{zhu2024minivln}  demonstrated strong performance through dataset scaling, we focus on comparing methods that use the same amount of data and do not directly compare with it.

\textbf{Method with LLMs/VLMs fine-tuning} These methods leverage pre-trained large foundation models by fine-tuning them on VLN-specific datasets, including NavCoT \cite{lin2024navcot}, LangNav \cite{pan2023langnav}, NaviLLM \cite{zheng2024towards}, and NavGPT-2 \cite{zhou2024navgpt2}. They typically use small or medium sized LLMs/VLMs such as LLaMA-7B \cite{touvron2023llama}, Vicuna-7B, Vicuna-13B \cite{vicuna2023}, and FlanT5 \cite{chung2024scaling}. Fine-tuning these models is computationally intensive, as discussed in Section~\ref{cost}.

\textbf{Method with pre-trained LLMs/VLMs} These methods leverage the zero-shot capability of pre-trained LLMs such as GPT-4 \cite{achiam2023gpt}, reducing computational costs. The methods include NavGPT \cite{zhou2024navgpt}, MapGPT \cite{chen2024mapgpt}, and DiscussNav \cite{long2024discuss}.

\subsubsection{Evaluation Metrics}
Following prior work \cite{anderson2018evaluation,chen2022think,zhou2024navgpt2}, we adopt a comprehensive set of navigation metrics, including Trajectory Length (TL), measuring average path length; Navigation Error (NE), the average distance to the target; Success Rate (SR), the frequency of task completion; Success weighted by Path Length (SPL), which accounts for path efficiency; and Oracle Success Rate (OSR), the success rate under an ideal stop policy.

\subsubsection{Parameters and Environment} In the pre-training stage, we use the pre-trained LXMERT \cite{tan2019lxmert} as initialization and train our model with a batch size of 32 for 100k iterations. Then we fine-tune the policy with the batch size of 8 for 20k iterations. We select the checkpoint with the highest SPL on validation unseen data and evaluate on other matrices. For image features, we use CLIP ViT-B/16 \cite{radford2021learning} to extract the features, and we use a single-layer feed-forward network for the Partial Contrast Module. We set the $\gamma$ and $\alpha$ in Equation~\ref{eq:total_loss} to be 1 and 0.5 respectively, and the object portion ration $\lambda$ equal to 0.2. For the parameter $m$, which is a threshold parameter to construct the sample selection matrix $A$, we set it equal to 1 for all experiments. We highlight the main settings here and adhere to the configurations outlined in previous work \cite{chen2022think} for any unspecified details.

\begin{figure}
  \centering
  \begin{subfigure}{0.23\textwidth}
        \includegraphics[width=\textwidth]{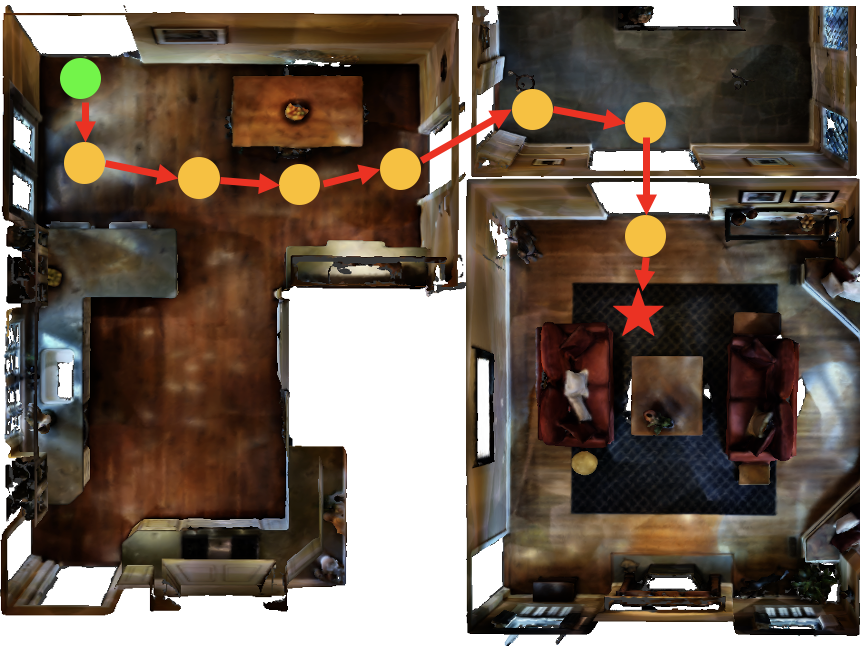}
        \caption{}
        \label{fig:case_eg_1}
    \end{subfigure}
    \hfill
    \begin{subfigure}{0.23\textwidth}
        \includegraphics[width=\textwidth]{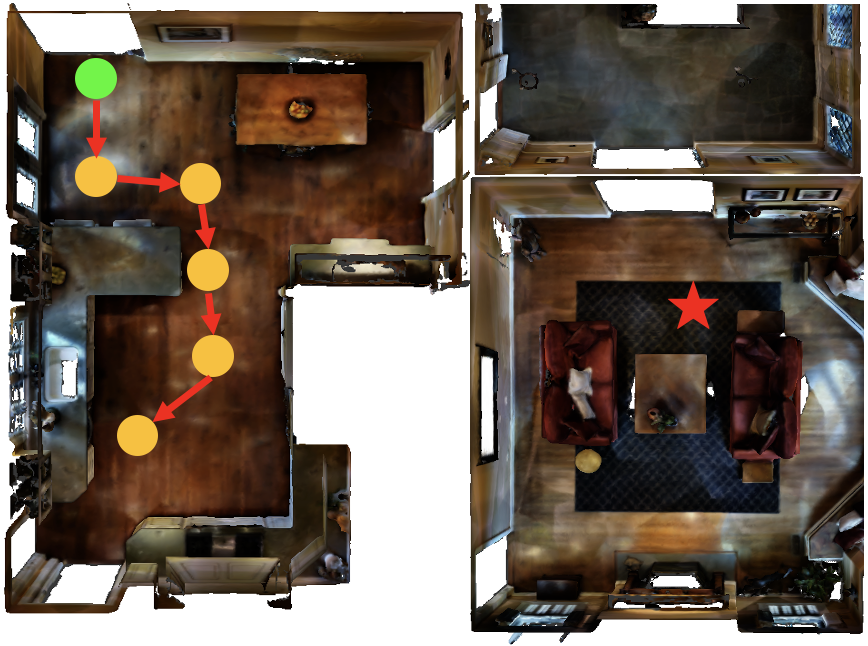}
        \caption{}
        \label{fig:case_eg_2}
    \end{subfigure}
  \caption{Case study on R2R with the instruction: \textit{Walk through the kitchen and into the living area with two red sofas facing each other. Stop just when you reach the black rug.} (a) Our method successfully completes the task. (b) DUET deviates in the wrong direction and fails to complete the task.}
  \label{fig:case_eg}
  \vspace{-10pt}
\end{figure}

\subsection{Overall Results}
\label{exp_rst}
The results of our method are shown in Table~\ref{tab:rst_r2r}, Table~\ref{tab:rst_reverie}, and Table~\ref{tab:rst_soon} for R2R, REVERIE, and SOON, respectively. Note that the baselines differ across datasets due to the unavailability of results for certain methods. Overall, our method consistently outperforms the baselines, demonstrating its effectiveness and robustness.

Specifically, on REVERIE and SOON datasets, our method outperforms the baseline methods by a considerable margin across all metrics. On R2R, our method surpasses other methods across most metrics. In particular, when compared with other peer methods lying in the category \textit{with pre-trained LLMs/VLMs}, our method outperforms existing methods significantly. While the performance gap is smaller compared to methods that fine-tune LLMs, our approach requires significantly fewer computational resources, as we leverage the zero-shot capabilities of VLMs without any fine-tuning.

We also coducted case studies and Figure~\ref{fig:case_eg} illustrates an example from R2R. Given the instruction, our method successfully reaches the correct destination (Figure~\ref{fig:case_eg_1}), while DUET deviates and fails (Figure~\ref{fig:case_eg_2}). This highlights the effectiveness of our approach.

\begin{table}[t]
  \centering
  \small
  \caption{Results on REVERIE. Our method demonstrates superior performance compared to existing approaches.}
  \setlength{\tabcolsep}{4pt}
  \renewcommand{\arraystretch}{1}
  \begin{tabular}{l | c c c | c c c}
    \toprule
    & \multicolumn{3}{c}{Val Seen} & \multicolumn{3}{c}{Val Unseen} \\
    & OSR $\uparrow$ & SR $\uparrow$ & SPL $\uparrow$ & OSR $\uparrow$ & SR $\uparrow$ & SPL $\uparrow$ \\
    
    \midrule

    Seq2Seq \cite{anderson2018vision} & 35.70 & 29.59 & 24.01 & 8.07 & 4.20 & 2.84 \\
    RCM \cite{wang2019reinforced} & 29.44 & 23.33 & 21.82 & 14.23 & 9.29 & 6.97 \\
    Airbert \cite{guhur2021airbert} & 48.98 & 47.01 & 42.34 & 34.51 & 27.89 & 21.88 \\
    HAMT \cite{chen2021history} & 47.65 & 43.29 & 40.19 & 36.84 & 32.95 & 30.20 \\
    DUET \cite{chen2022think} & 73.86 & 71.75 & 63.94 & 51.07 & 46.98 & 33.73 \\
    GOAT \cite{wang2024vision} & - & 78.64 & \textbf{71.40} & - & 53.37 & 36.70 \\
    
    \midrule
    
    WPCL (Ours) & \textbf{75.31} & \textbf{79.41} & 71.39 & \textbf{53.94} & \textbf{54.03} & \textbf{37.41} \\
    
    \bottomrule
  \end{tabular}
  \label{tab:rst_reverie}
\end{table}

    
    
    
    
    

\begin{table}
  \centering
  \small
  \caption{Results on SOON. Our method demonstrates superior performance compared to existing approaches.}
  \setlength{\tabcolsep}{4pt}
  \renewcommand{\arraystretch}{1}
  \begin{tabular}{l | c c c | c c c }
    \toprule
    & \multicolumn{3}{c}{Val Seen} & \multicolumn{3}{c}{Val Unseen} \\
    & OSR $\uparrow$ & SR $\uparrow$ & SPL $\uparrow$ & OSR $\uparrow$ & SR $\uparrow$ & SPL $\uparrow$ \\
    
    \midrule   
    
    GBE \cite{zhu2021soon} & 28.54 & 19.52 & 13.34 & 21.45 & 12.90 & 9.23 \\
    DUET \cite{chen2022think} & 50.91 & 36.28 & 22.58 & 43.00 & 33.44 & 21.42 \\
    GOAT \cite{wang2024vision} & 54.69 & 40.35 & 28.05 & 50.63 & 40.50 & 25.18 \\
         
    \midrule
    
    WPCL(Ours) & \textbf{55.13} & \textbf{41.24} & \textbf{28.53} & \textbf{52.21} & \textbf{42.92} & \textbf{26.57} \\
    
    \bottomrule
  \end{tabular}
  \label{tab:rst_soon}
\end{table}

\subsection{Further Analysis}
\label{exp_sensitivity}

\subsubsection{Sensitivity to Prompt} 
We introduced the prompt used to generate the supervisory signal in Section~\ref{method_supervised_vlm}. Here, we further analyze the effect of different prompt designs on extracting object information from images.

Specifically, we found that specifying criteria for desired objects is crucial. Without such specificity, the VLM often identifies irrelevant objects like floors, rooftops, or grass. While these elements are indeed present in the image, they are ubiquitous, and including them would classify all observations as positive samples, thus undermining the effectiveness of our algorithm. Therefore, we ask the model to \textit{Ignore the general environmental objects such as walls, floors, or rooftops}.

We also observed that specifying the expected output format is essential. Without clear instructions, the VLM may produce outputs in an inconsistent format, complicating post-processing. To mitigate this, we included the directive \textit{Please only answer with a list of objects, and separate each object by a comma} in the prompt, ensuring the output is structured and proper for subsequent use.

\subsubsection{Sensitivity to VLM choice}
We use the pre-trained LLAVA 1.5 \cite{liu2023llava} to obtain the supervisory signal in our primary experiments, given its proven effectiveness and efficiency with a moderate parameter size. To investigate the impact of the VLM selection, we conducted further experiments using LLAVA 1.6 \cite{liu2024llavanext}, OpenFlamingo \cite{awadalla2023openflamingo}, and BLIP-2 \cite{li2023blip}, applying the same prompt across all models. We observe the outputs from these models were largely consistent.

\begin{figure}
    \centering
    \includegraphics[width=0.9\linewidth]{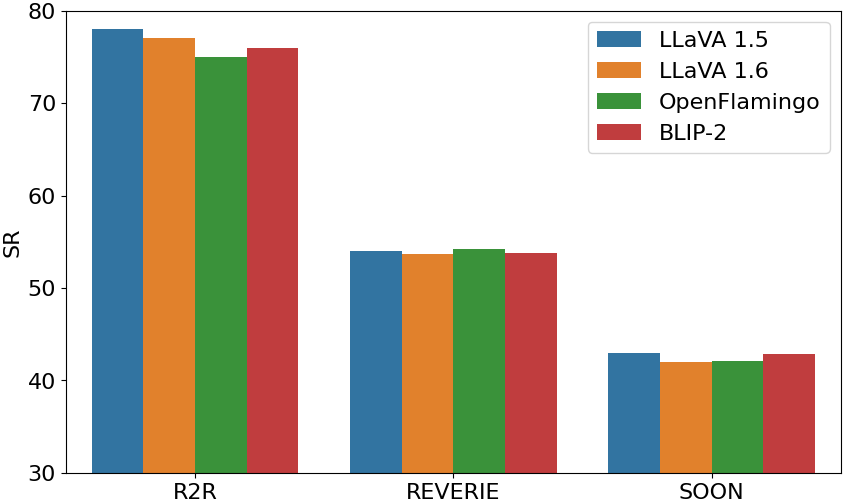}
    \caption{Sensitivity to VLM choice. We report the Success Rate (SR) on the validation unseen set for all tasks, using various Vision-Language Models (VLMs). The results indicate that the choice of VLM has minimal impact on performance, highlighting the robustness of our method.}
    \label{fig:sensitivity_llm}
    \vspace{-10pt}
\end{figure}

We visualize the effect of these models on the agent's overall performance in Figure~\ref{fig:sensitivity_llm}, where we observe minimal differences in performance. We hypothesize that this is due to the relative simplicity of object recognition tasks for VLMs, allowing all models to perform similarly well.

This finding underscores the robustness of our method, which minimizes reliance on specific VLMs. Even though these models may exhibit significant variation on other tasks, our approach remains generalizable and effective across different foundation models.

\subsubsection{Accuracy of VLM object detection}
Based on our case studies such as the example in Figure~\ref{fig:partial_contrast}, we find that detection is highly accurate, achieving near-perfect quality. However, due to the absence of object-level labels in the datasets, we cannot directly evaluate VLM detection accuracy. As analyzed earlier, different VLMs perform similarly well, reinforcing their proven object detection capability \cite{zhang2024vision}.

\subsection{Compuatational Cost}
\label{cost}
Our method shows computationally efficiency as it does not require VLM fine-tuning. It runs on a single 24GB NVIDIA TITAN RTX GPU, while existing methods require substantially more resources for fine-tuning. For instance, NavCOT \cite{lin2024navcot} uses a 32GB V100, NavGPT-2 \cite{zhou2024navgpt2} an 80GB A100, LangNav \cite{pan2023langnav} 72 V100 GPUs, and NaviLLM \cite{zheng2024towards} 8 A100 GPUs. This underscores the efficiency of our approach, which leverages VLMs only for inference.






\section{Related Work}
Vision-and-Language Navigation (VLN) tasks \cite{anderson2018vision,qi2020reverie} aim to enable agents to follow human instructions to reach specific destinations. To enhance action planning, existing methods often incorporate techniques such as historical state tracking \cite{kareer2023vinl,wasserman2024exploitation}, self-correction mechanisms \cite{ma2019regretful,ke2019tactical}, navigation map generation \cite{chen2022think,liu2023bird,wang2023gridmm, huang2023visual,xu2024robot,suomela2024placenav} and the use of external knowledge prompts \cite{li2023kerm,zhao2023zero,lyons2023wavn,guan2024loc,xu2024aligning}. To better align visual and language representations, some approaches leverage general visual-linguistic representations built from generic frameworks \cite{chen2022think,guhur2021airbert,majumdar2020improving}, while others rely on additional supervision through data augmentation \cite{li2024panogen,parvaneh2020counterfactual,tan2019learning,wang2022counterfactual,wang2023scaling} and optimized training strategies \cite{huang2019transferable,ma2019self,wang2018look,wang2019reinforced,zhu2020vision}.

Recently, with the rise of pre-trained Large Language Models (LLMs) and Vision-Language Models (VLMs), several studies have explored how to integrate this pre-trained knowledge into VLN. Some works such as NavGPT \cite{zhou2024navgpt}, DiscussNav \cite{long2024discuss}, MapGPT \cite{chen2024mapgpt}, PixNav \cite{cai2024bridging} showcase the potential of off-the-shelf LLMs for navigation by using its zero-shot capability. However, despite the use of advanced models such as GPT-4 \cite{achiam2023gpt}, these approaches still exhibit a performance gap compared to supervised methods. To address this, methods like LangNav \cite{pan2023langnav}, NavCoT \cite{lin2024navcot} and NaviLLM \cite{zheng2024towards} fine-tune LLaMA-7B \cite{touvron2023llama} on VLN-specific datasets to assess the role of language as a perceptual tool for navigation.


\section{Conclusion}
In this paper, we propose the Weakly-supervised Partial Contrastive Learning (WPCL) framework for Visual Language Navigation, which effectively integrates pre-trained Vision-Language Models (VLMs) into the perception process of agents without fine-tuning it, thereby maintaining computational efficiency while reducing reliance on potentially biased outputs. Through causal analysis and tailored partial contrastive learning, our approach achieved strong performances across multiple benchmarks, highlighting its potential to advance VLN, offering a balanced solution that leverages pre-trained models' strengths without their associated limitations.





\bibliographystyle{IEEETran}
\bibliography{reference}

\end{document}